**PREPRINT – FINAL MANUSCRIPT PUBLISHED IN NATURE MACHINE INTELLIGENGE**

https://www.nature.com/articles/s42256-025-01050-6

-------------------------------------------------------------

# Which AI for robotics? Challenges set forth to guarantee safe, ethical and sustainable deployment of robots for and with humans

**Aude Billard\*,**
Learning Algorithms and Systems Laboratory (LASA), École Polytechnique Fédérale de Lausanne (EPFL), Lausanne, Switzerland

**Alin Albu-Schaeffer**,
Institute of Robotics and Mechatronics, DLR-German Aerospace Center/Department of Informatics, Technical University of Munich;

**Michael Beetz**,
Institute for Artificial Intelligence, Computer Science Department, University of Bremen;

**Wolfram Burgard**,
Department of Engineering, University of Technology Nuremberg, Germany

**Peter Corke**,
Center for Robotics, Queensland University of Technology, Brisbane, Australia

**Matei Ciocarlie,**
Columbia University, New York City, USA

**Ravinder Dahiya**,
Northeastern University, Boston, USA.

**Danica Kragic**,
School of Computer Science and Communication, Royal Institute of Technology, Stockholm, Sweden

**Ken Goldberg**,
University of California, Berkeley, USA

**Yukie Nagai**,
International Research Center for Neurointelligence, The University of Tokyo

**Davide Scaramuzza**,
Robotics and Perception Group, University of Zurich, Switzerland

\*Corresponding author. E-mail: aude.billard@epfl.ch

# Which AI for robotics? Challenges set forth to guarantee safe, ethical and sustainable deployment of robots for and with humans

AI technologies, including deep learning, large-language models have gone from one breakthrough to the other. As a result, we are witnessing growing excitement in robotics at the prospect of leveraging the potential of AI to tackle some of the outstanding barriers to the full deployment of robots in our daily lives. However, action and sensing in the physical world pose greater and different challenges than analysing data in isolation. As the development and application of AI in robotic products advances, it is important to reflect on which technologies, among the vast array of network architectures and learning models now available in the AI field, are most likely to be successfully applied to robots; how they can be adapted to specific robot designs, tasks, environments; which challenges must be overcome. This article offers an assessment of what AI for robotics has achieved since the 1990s and proposes a short- and medium-term research roadmap listing challenges and promises. These range from keeping up-to-date large datasets, representatives of a diversity of tasks robots may have to perform, and of environments they may encounter, to designing AI algorithms tailored specifically to robotics problems but generic enough to apply to a wide range of applications and transfer easily to a variety of robotic platforms. For robots to collaborate effectively with humans, they must predict human behavior without relying on bias-based profiling. Explainability and transparency in AI-driven robot control are not optional but essential for building trust, preventing misuse, and attributing responsibility in accidents. We close on what we view as the primary long-term challenges, that is, to design robots capable of lifelong learning, while guaranteeing safe deployment and usage, and sustainable computational costs.

The last decade has witnessed impressive advancements in the development and practical application of Artificial Intelligence (AI) technologies, in particular for systems based on Deep Learning (DL) over multi-layer artificial neural networks (ANNs). Though ANNs are not recent concepts, several factors have contributed to a fast-paced acceleration in their performance and scalability. On one side, computing platforms, such as Graphical Processing Units (GPUs), have become available, offering increased computational power and allowing to create "deeper" networks (i.e. with more hidden layers). On the other hand, the exponential growth of multimodal, digital information, and open-source software available on the Internet has made vast amounts of data easily available for the creation of training and test datasets.

The first demonstration of the potential of these technologies came in the early 2010s, when deep networks started overcoming previous systems in visual recognition challenges [1]. Since then, there have been important applications of these systems on several different computational tasks.

Great expectations currently surround the applications of new AI systems to robotics. Once again, this is not a novel concept, because learning algorithms have been used to control robots for decades. But there is hope that the current fast-paced scaling-up of AI's performances may translate into a similar scaling-up of robotic capabilities and help solve some long-standing challenges that have so far limited robots' autonomy in challenging environments or their capability to interact effectively and safely with humans. For example, the classic control-theoretic and state estimation methods for robots, that were developed for controlled and structured industrial environments, struggle to adapt to the high complexity and intrinsic unpredictability of outdoor natural environments, or even to the diversity of objects that can be encountered in a typical home. It is tempting to expect that advancements on these problems will mirror what happened for Go – a boardgame that was famously impossible for existing classic computer programs to master mathematically. But deep Learning came and vastly surpassed human abilities after playing billions of games with itself.

We cannot expect that approaches that excel in purely data and software-based gaming environments, or image or text generation, be readily applied to real-time sensing, planning, control, and navigation for physical machines operating in complex, unpredictable environments — especially those involving human interaction. Action and sensing in the physical world pose different, and arguably greater challenges than playing games: the state space is bigger, training data are not readily available nor easily generated, and safety and reliability are non-negotiable. It is paramount to determine when AI is needed and when not and to identify which technologies, among the vast array of architectures and learning models now available in the AI field, can be successfully applied to robots; how they can be adapted to specific robot designs, tasks, environments.

Next, we provide a brief review of main AI for robotics achievements since the 1990s and propose a short- and medium-term research roadmap listing its promises and challenges.

**A brief historical review:** Allowing robots to operate autonomously in novel situations and to approximate the dexterity and agility of living organisms have been key challenges for robotics since at least the 1960s [2] [3] [4]. For several decades, robotics researchers have been experimenting with AI as a potential solution to those challenges, and there is now a sizable literature on how to leverage these techniques to tackle robotics problems that had previously proven hard to solve. These studies have provided insight into which tasks are more amenable to be learned rather than formally programmed.

Among the vast array of AI techniques used in robotics since the 1990s, two principal types of algorithms and data gathering stand out. On one side there is a family of algorithms that allow robots to learn from expert data, typically provided by a human demonstrator who performs the target action while their movement is captured by visual or motion sensors. Called alternatively Programming by Demonstration, Learning from Demonstration (LfD) or Imitation Learning, this approach has proved applicable in tasks ranging from grasping to manipulation of complex objects [5] [6] [7]. LfD algorithms could produce impressive results, such as catching objects in flight or control complex flying manoeuvres [8] [9], while relying on small datasets. The main limitation of LfD has historically been the intrinsic need to have a human operator with a good knowledge of the task available for training the robot, often across many training sessions. To address these challenges, current efforts are directed to learning from non-experts or suboptimal demonstrations, or from large collections of human and robot actions [10] [11] [12]. Other approaches, such as active learning [13], one-shot and behavioral imitation [14] or behavioral cloning [15] [16], have also been proposed as a way to improve the efficiency of LfD: these techniques allow the robot to query the expert for demonstrations only when required, to learn a complete behavior from a single demonstration, or to start by acquiring experience in a self-supervised fashion and then use this experience to develop a model which is then used to facilitate learning of particular task by observing an expert. All of these have been shown to require fewer post-demonstration environment interactions than other techniques.

The other type of learning algorithms, known as reinforcement learning (RL) [17], enables robotic systems to learn through trial and error without a prior formalization of what constitutes the correct control policy. RL is commonly deployed on computer simulations of the robots and its environment to create enough learning cycles and learn a robust enough policy before testing it on the actual robot, to address challenges with the feedback from real robots which may be incomplete or delayed. Although RL balances exploration and exploitation, the exploration phase remained often time intensive and did not easily scale to high dimensions. Recent advances leverage the increasing effectiveness of large-data learning, achieving notable success in applications such as locomotion for legged robots – both quadrupeds and humanoids – as well as flying robots [18] [19] [20]. Despite vast progresses done in the design of realistic simulators, transfer of learning conducted in simulation, far from the complexity of the real world to real environments, known as the sim-to-real problem, remains a challenge [21]. In addition, IR success depends on a good prior knowledge of how to define an effective reward metric and assess the robot's performance against it.

Some of these challenges can be resolved when using LfD and RL in combination to leverage the strength of both techniques while mitigating their limitations. LfD can be used, for example, to reduce the search space in RL by bootstrapping it with good examples [22], reducing training time of large models [23], or to infer the reward and the optimal control policy simultaneously, a technique known as Inverse RL [24]

**Potential for novel applications and commercial deployments:**

Many advances initiated in academic research have found their way to commercial applications. AI-powered robots that can pick and sort packages of various sizes are increasingly deployed in e-commerce warehouses. Learning enables online adaptation in tasks like pick-and-place on assembly lines, which were once rigidly pre-programmed. Robots can now adjust trajectories if an object is misplaced, or its shape or weight is unexpected. Autonomous driving, which started in the early 2000s, is now available - ranging from partial autonomy in most models currently on the market to pilots of full autonomy underway, in limited situations and certain cities.

While AI is now pervasive in all areas of robotics, an area of application of particular interest is the field of soft robotics, where the deformable, continuum nature of robot bodies and their complex interaction with environments makes the processing of sensor data, state estimation, and control particularly challenging. Soft robotics natural compliance may ease the usage of robots in areas requiring direct interaction with humans and address global issues through biodegradable solutions. AI may offer an alternative to or complement rigid mechanics control approaches, notably to process soft robots' nonlinear, hysteresis-prone and heterogeneous sensor data stream [25]. A notable example is the recent application of convolutional neural networks to interpret the wealth of data streaming from a soft glove's artificial skin, enabling real-time recognition and control of grasps on objects [26].

**Short- and medium-term challenges:**

We have only begun to scratch the surface of the potential of RL, LfD, and other flavors of AI for robotics. In the short- and long-term, many challenges lay ahead - ranging from software and hardware developments to theoretical and algorithmic advances (see Figure 1).

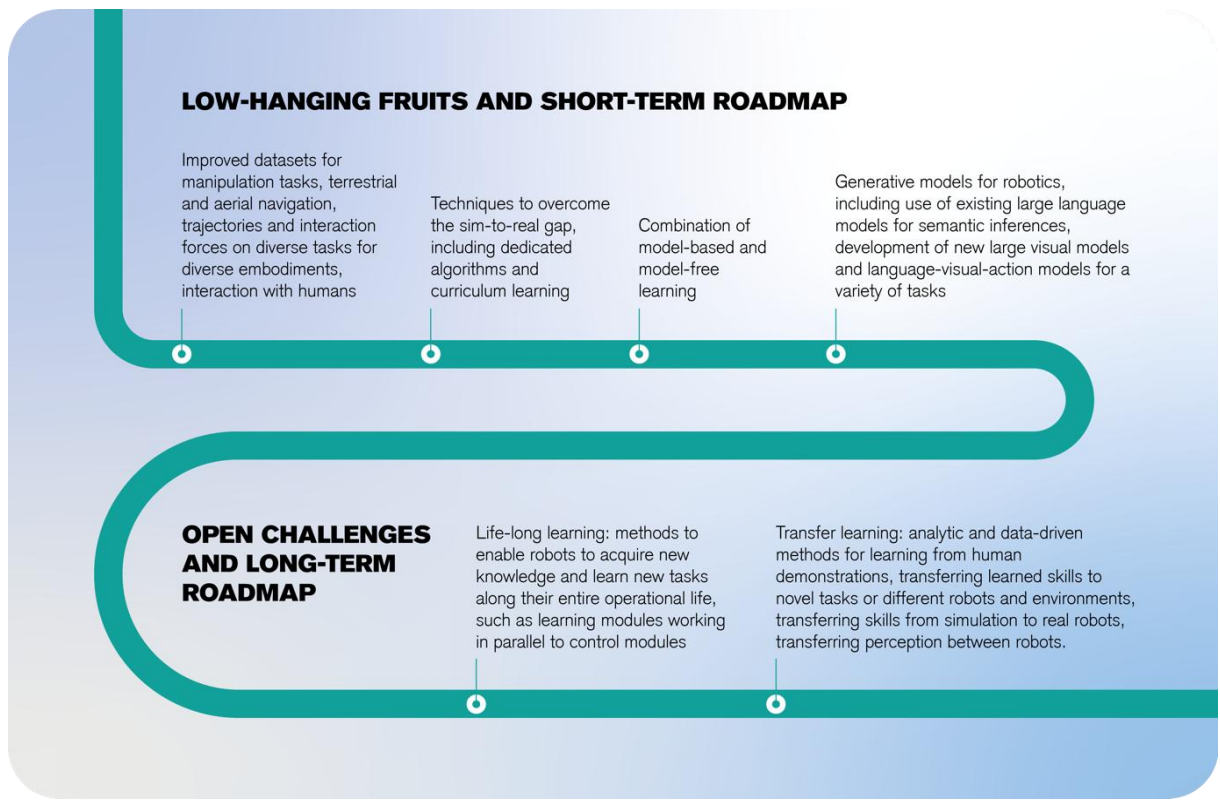

FIGURE 1: *Short-term and long-term challenges for further and long-term endorsement of AI in robotics,* enabling robots to learn continuously, autonomously and when interacting with humans and other robots. *Challenges are ordered by increasing level of complexity but may not be resolved sequentially. Rather, research proceeds in parallel along many of these directions.*

<u>*Creating and maintaining representative datasets*</u>: An intrinsic limitation in robot learning as compared to other AI application domains is that there are no ready-to-use and easily available large datasets that can be used to train ANNs on sensing and control tasks, comparable to the vast repertoire of images and text that could be downloaded from the Internet and used to train image recognition or text generation algorithms. Generating *ex novo* enough iterations of a robotic task to train an ANN can be exceedingly costly and time consuming, or simply impossible. Too many robots would be destroyed during failed attempts at a task, and in some cases (such as autonomous flying robots) this would create risks for humans.

For some tasks, reference databases can be created but require an organized and multi-centric effort. For example, in the case of visual imitation learning, attempts are being made at creating an analogue of ImageNet for grasping and manipulation. For instance, the Dexterity Network (Dex-Net) research project develops code, datasets, and algorithms for generating parallel-jaw robot grasps and metrics of grasp robustness based on physics for thousands of 3D object models and supports researchers in finding robust grasps and training machine-learning models to generate a wealth of grasping strategies, notably resolving the so-called "bin-picking" problem, namely picking many unfamiliar objects stacked in arbitrary orientations, that had for long been a benchmark challenge in the field [27].

Large datasets are also being created for terrestrial navigation tasks, thanks to cars now collecting large amounts of images routinely, from professional mapping services such as Google Maps to dashcams becoming increasingly common on private vehicles. These databases are typically available to companies on a proprietary basis, but if privacy and IP issues can be dealt with, it is foreseeable that some of them can become available to researchers. The challenge is bigger for aerial navigation, because of the many different perspectives from which a drone can observe the same scene, at vastly different altitudes and tilting orientations with respect to the ground. Curated datasets that reconcile these various viewpoints will be required.

Beyond visual data, robot learning needs datasets of robot actions in the form of trajectories and interaction force profiles associated with various tasks. Datasets on specific robot bodies and tasks do exist, but they are typically too narrow for large-scale machine learning. Combining datasets from diverse embodiments and on diverse robotic tasks can be a solution to reach the required scale. For example, an effort has recently been launched to combine several datasets on robotic manipulation, each one based on a specific robot body and skills and has provided a proof of concept that such a combined dataset could be used to train a policy for a given task more effectively than by using a dataset specific to that task [28].

Possibly the biggest challenges in terms of dataset creation are related to close interaction with humans, as the complexity and variability of both physical interactions and communication with humans and the need for enhanced safety guarantees currently prevents the rapid creation of datasets either through real experiments or in simulation. Ethical issues also put strict limits on what data can be collected and stored about human subjects and how they can be labelled, for example, by ensuring that subjects are not recognizable, that no sensitive information about them can be inferred from the data, or that images of a human subject cannot be reused in a different context, including being used for different training objectives than initially specified. An additional complication is that robots and humans perceive the world and interact with it in very different ways: while humans rely on multimodal information combining visual, acoustic, and haptic information, robots mostly rely on vision or on other bands of the electromagnetic spectrum, and while they can see more than humans do (including in low light or through obstacles) they remain incapable of analysing complex visual scenes. Building factories of human demonstrators can allow to benchmark several well-defined tasks but will be insufficient to account for the vast complexity of human activities in homes.

*From simulation to reality and back:* Simulations offer a partial solution when it is not possible to create a large enough dataset. Several robotic simulators are available to the robotic community (examples include Algoryx, Bullet, Gazebo, Isaac Sim, MuJoCo, RoboDK, Genesis) and have been used for a long time to test and improve classic model-based control algorithms before applying them to real robots. The accuracy of their physics-engines has greatly improved, also thanks to their commercial use in computer gaming. Reliable physics-based simulators can, for example, simulate locomotion on complex terrains and manipulation on realistic objects in home environments, reducing the time needed for training by enabling searches over several thousands of iterations before transfer to the real world [29] [30] [31].

However, overcoming the sim-to-real gap, i.e. the discrepancy between the robot's performance in the real world and in the simulated environment, remains a challenge. This gap can be the result of multiple factors: the simulator's model can be exceedingly simplified with respect to the actual physical robot; the variability of the environmental conditions can be too large to be captured by a model; the physics simulator can fail to accurately capture the physics of the real world, especially when it comes to contact forces and deformable surfaces. There are many techniques to overcome the sim-to-real gap. A small amount of data from the real world can be collected and used to increase the realism of the simulator [32], to achieve online real-time adaptation of quadruped locomotion to changing terrains, payloads, wear and tear [33].

Closing the loop from reality to simulation holds significant potential, yet, bridging the real-to-sim gap by modifying simulators using real-world data has received much less attention than the reverse.

*Leveraging large generative models for robotics:* Much of the current excitement around AI focuses on generative AI, and Large Language Models (LLMs). They are mostly based on the "transformer" deep learning model, which around 2017 emerged as an alternative to both recurrent and convolutional neural networks, allowing the speedup of learning (in particular of textual information) by processing information sequences in parallel [34]. Similar principles may lead to step changes in robotics. The question is which structure (inductive bias) should we embed in robot learning algorithms to enable similar step changes in the control of robots?

LLMs are attractive for robotics on multiple levels. Existing LLMs can be adapted to support human-robot interaction based on natural language, essentially making it easier to control a robot through written or verbal instructions, in any human language, and allowing them to respond to humans accordingly. Attempts are also being made at using LLMs in robot navigation in new and unfamiliar environments, to support semantic guesswork, essentially using their inferences [35].

Another family of generative models are language-vision models, that are trained on text/image pairs or annotated videos found on the Internet, and that can be used to generate synthetic images and videos from text prompts [36]. These models can also be applied to robotics, for example, to improve object recognition in manipulation and navigation tasks, and allow tasks to be specified in terms of what can be seen by the robot. A new generation of large visual models can be purposely built for robotics, trained not (or not exclusively) on text/image pairs from the Internet, but on navigation datasets such as those described in the previous section, produced by cameras during actual navigation in real environments. A first step could be learning to generate expectations on domestic spaces, i.e. using datasets of images of homes and offices or information from sensorised objects to generate reliable predictions on what a robot moving around such an environment may encounter. The same approach could then be extended to terrestrial and aerial navigation, creating models that can understand and contextualize visual information and incorporate a model of the robot's own physics and behavior to predict what it will see next.

The most recent developments in the field are language-vision-action models that add action to the equation. Examples of such models are being proposed, trained by fine-tuning vision-language models with both Internet-scale visual-language tasks and robotic trajectory data. By expressing the robot actions as text tokens and incorporating them into the training set together with natural language tokens, these models can learn to output robot actions like LLMs output text [37]. Initial results are encouraging, but the challenge of feeding such models with suitable datasets (see section 2.1), effectively mapping vision to action, and providing the system with the reasoning capability to correctly anticipate the consequences of its actions, will have to be a core research focus for several years. Another challenge is to verify the logic and feasibility of the plan generated by LLMs, an issue that is well addressed in logic-based planning [38].

*Prior knowledge and combining AI with control methods*: For physical robotics, incorporating prior knowledge on both robot and environment dynamics in combination with control methods with provable guarantees, is a more sensible way forward than a totally bottom-up, knowledge-agnostic approach to learning. In aerial robotics, for example, neither learning nor aerodynamics-based control alone can help solve the challenge of approximating the agility of birds' flight: coupling sensing and perception with the full body dynamic, allowing a drone to have instant reactions in flight and cancel perturbations, or on the contrary profit from the wind, efficiently combining flapping of wings and gliding (in the case of a winged drone) to save energy. These challenges will require a combination of learning for building improved aerodynamics models with control methods for guaranteeing flight stability.

Another reason for combining models and formal knowledge with machine learning is that a system only based on the latter would be prone to failures that can neither be predicted beforehand nor fully explained afterward, as exemplified by "hallucinations" observed in LLMs. Many current deep learning (DL) models are intrinsically non-explainable, a problem that becomes even more critical when AI is applied to robots. Since most future robots are expected to operate in safety-critical scenarios such as autonomous navigation or close interaction with humans, no regulatory agency would approve their use unless their behavior can be predicted, and performance guarantees met – failures must be explained and corrected which is currently not feasible with model-free deep learning. This is a serious limitation to almost all applications where harm to humans is possible, for example, in the medical field, aeronautics, logistics and transportation, and domestic use.

AI-powered robots will need models of the actions that they are about to do, and these models must be explicitly represented to reason about the consequences. For example, a robot designed to work in a chemical lab, whose task is to pour chemicals into different containers, needs to know what happens when an acid is mixed with a base. Whenever a human comes into play, the robot needs an actual theory of mind modelling and excellent knowledge of what the human may do and how the human might interpret the robot's task. Such human modelling can quickly become more complicated than the model of the robot itself [39].

Numerous efforts are directed to merge control theory and machine learning, proving that this can ultimately speed up learning, increase the robustness of the learned model, and enhance its safety [40] [41]. For example, a standard machine learning algorithm optimization can be modified to encompass penalties for violations of theoretical constraints guaranteeing convergence or stability, or numerical bounds to enforce estimates of plausible values for physical quantities such as stiffness and mass. [42] In a similar vein, training of deep RL can be guaranteed to generate stable trajectories [43], or be enhanced by incorporating reference motions generated through control model, covering a broad range of velocities and gaits [44], serving as targets for the RL policy to imitate. Control theory and DL have also been combined to optimize grasps, using DL to find an initial policy that is then refined with model-based algorithms, thus sizeably speeding up computing [45].

**Long-term challenges**

The most exciting, but also most challenging, long-term promise of AI for robotics is to enable robots to continuously acquire new knowledge, a dream dating back the 90's [46]. It requires two ingredients, which we discuss next.

*Life-long learning:* If the goal of robot learning is to approximate the way living organisms – humans included – learn tasks, then future robots will need to be able to acquire new knowledge and learn new tasks along their entire operational life, instead of relying on an initial training dataset that could never prepare them for the complexity and variability of the real world.

Endowing robots with the ability to learn continuously poses huge technical and regulatory challenges. Lifelong learning requires new paradigms based on incremental learning that can convert input-output learning into structured knowledge, combining the power of learning with the paradigms of expert systems. It requires learning modules, and suitable hardware with required compute and memory, working around the clock on the robot in parallel to the control module enacting the policies that were already validated. It brings along difficult questions, such as: how do we get some assurance about the performance of the system? How can we test the system, provided we can't know in advance the situations it will encounter and how it will learn from them? How do we select the things the robot can forget to make room for learning new things? How do we make sure that whenever it learns something new it has not forgotten how to do something important that it could do yesterday? These problems will need to be investigated in close collaboration with neuroscientists and developmental psychologists, to understand how humans achieve continuous and diverse cognitive development transitioning from one task to another, how this mechanism can be reproduced in neural networks, and how they can be implemented in robots. These problems will also translate into major regulatory issues to verify that an evolving system maintains the safety and reliability standards requested for market certification as its capabilities change with new learning.

Possibly the main challenge for life-long robot learning will be to be able to scale up current learning methods. Many robots will not stay the same for their whole operational life. After five or eight years of operation, a robot may have to mount a different gripper, or a different motor. The objects it must manipulate and the environment in which it operates may also have changed. When that happens, the acquired knowledge that allows the robots to pick up and manage different objects may not automatically transfer to a slightly modified platform. We lack good algorithms to transfer knowledge automatically, without retraining or human intervention, across even small changes in the embodiment.

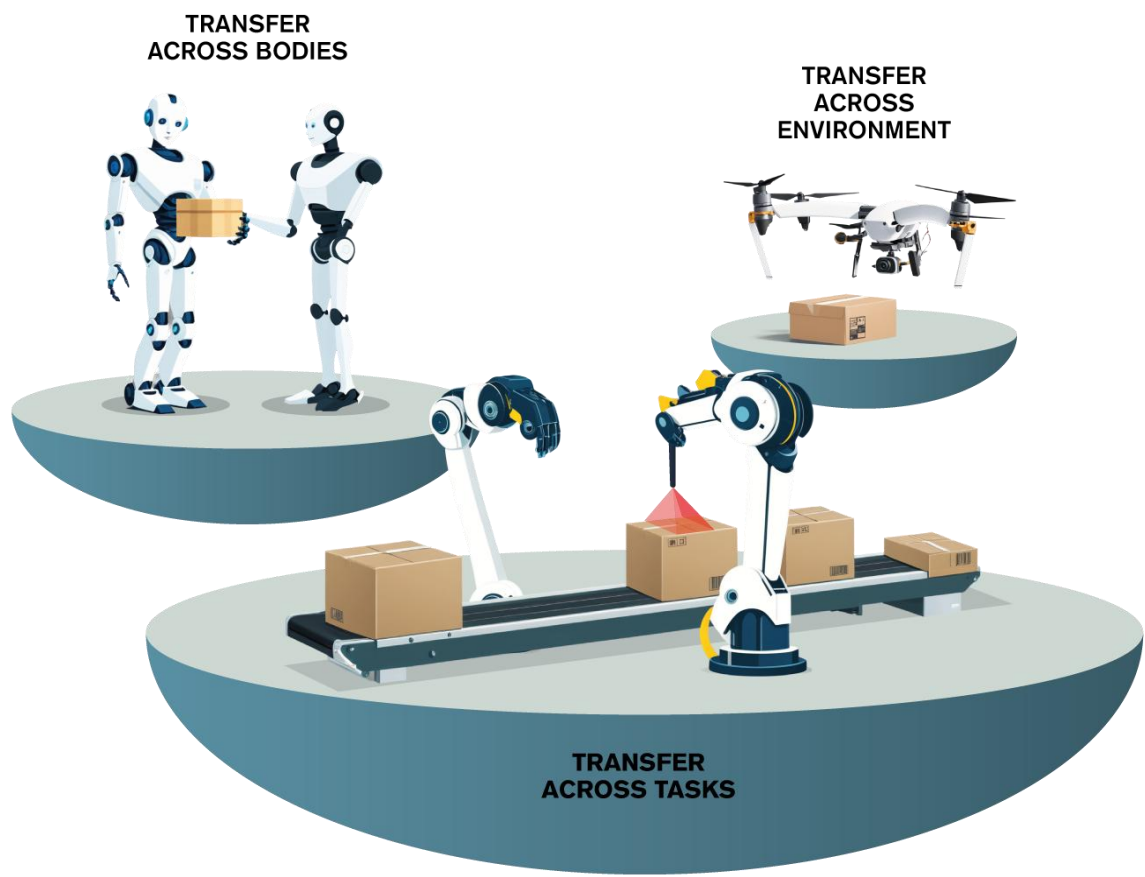

*Figure 2: The ability to transfer learning across robot bodies, tasks and environment is fundamental for achieving collaborations of different robots on one task.*

*Transfer learning:* Future robots will need to be able to *transfer* what they learn: from one task to another, from one environment to another, and from one robot to another (Fig.1). Human intelligence relies on the ability to apply the knowledge acquired in one domain to new domains - thus solving new problems and facing unexpected situations – and to share knowledge among individuals. Similarly, robots need analytic and data-driven methods for learning skills from human demonstrations, transferring learned skills to novel tasks or different robots and environments, transferring skills learned in simulation to real robots, transferring learned perception routines between robots.

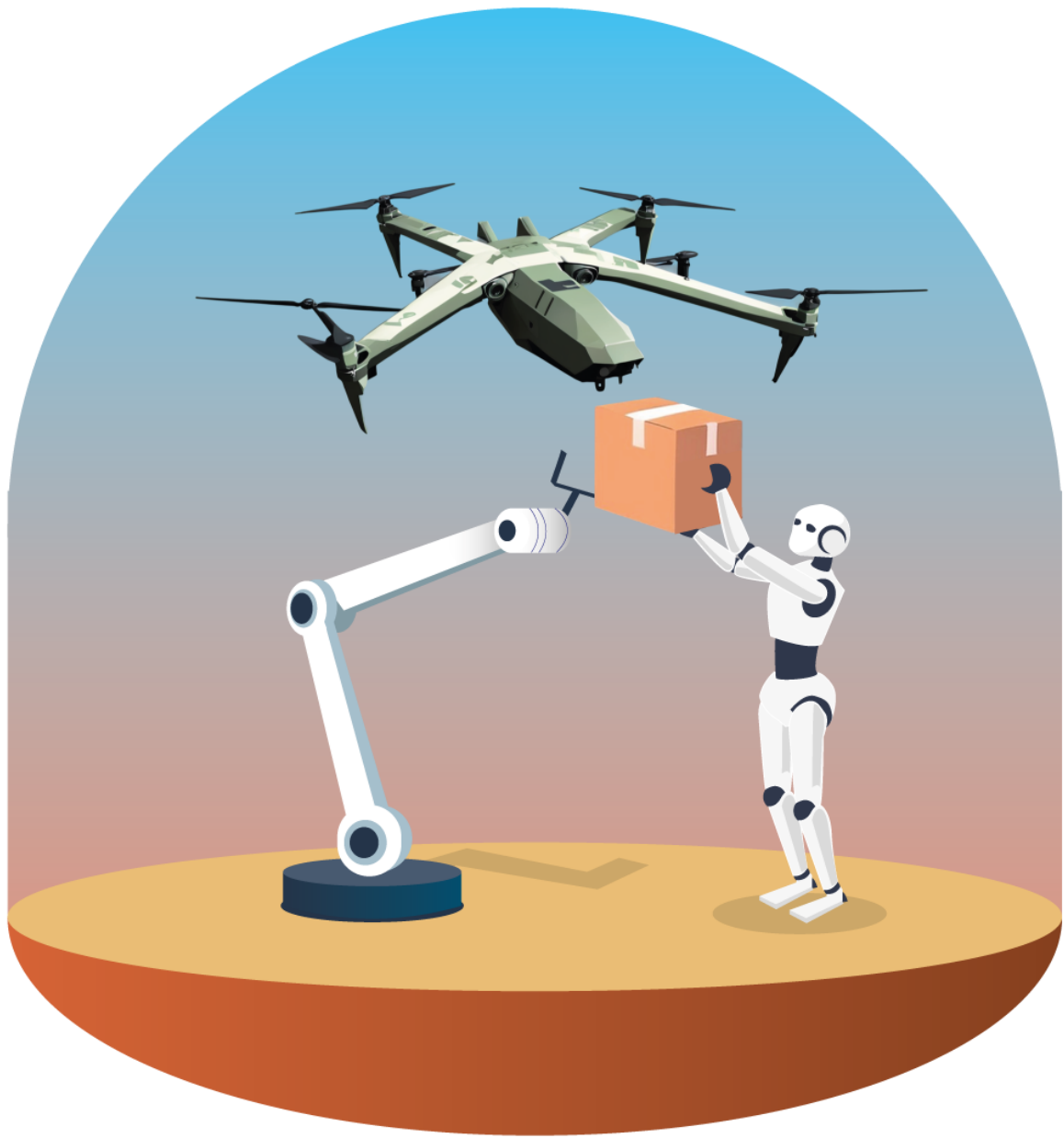

*Figure 3: Handing a package from a drone to a humanoid robot or single-arm robot manipulator requires to reconcile drastically different perception, from different viewpoints and sensors, and distinct robot actions from unimanual to bimanual actions (inspired by 2023 EuRobin Hackaton).*

Transferrable robot learning requires algorithm to answer three questions a) *what to transfer*? one must develop criteria to select which prior knowledge should be ignored and which should be transferred when dealing with new environments, objects, and tasks, b) *how to transfer*: one must devise algorithm capable of using the transferred knowledge, while filling in the gaps by conducting additional targeted search or by autonomously seeking additional knowledge on sensors, kinematics, actuators, electronic hardware specific to the new context, c) *when to transfer*; this requires developing algorithms to recognize similarities across environment, objects, tasks constraints, establishing if transfer of knowledge is at all possible in every specific case or if entirely new knowledge and novel learning cycles are needed.

*Safe exploration*: many important real-world applications are characterized by very high-dimensional state spaces, with incomplete observability, making live exploration essential considering sim-to-real limitations. We must invest in approaches by which robots can explore their environment in an effective and safe manner to both the robots and its surroundings. Such exploration is likely to be necessary throughout the lifetime of the robot, as discussed next.

**Looking Forward:** Deployment of AI and robotics at large has become a more tangible target, possibly foreseeable in the next decade. AI has the potential to expand largely the capabilities and range of applications of robotics, to tackle the many challenges posed by acting in real-time in a rapidly evolving world with only a partial understanding of their surroundings, and of humans.

Transfer learning is likely to hold one of the few keys to the fundamental issues of task and application scalability. But, for large scale deployment in applications ranging from manufacturing to logistics and homes, robots will need to learn new tasks quickly and effectively learn. A team of robotics experts with advanced degrees collecting weeks of data and then running hyperparameter tuning on the resulting policies for any new task the robot needs to learn is clearly not the way to go. Rather, robots capable of leveraging prior knowledge and complementing this through learning from punctual, minimal and intuitive guidance, e.g. through verbal instructions combined with visual demonstrations provided by domain experts who are not roboticists, are more likely to guarantee continuous and speed kill acquisition and adaptation.

Devising sparse and efficient use of AI and data will also be necessary, especially when data is gathered from high resolution sensing, such as when using artificial skins. A combination of hardware and software AI, such as "near sensor" computing where AI comes after the sensor data has been scaled down by distributed sensory hardware, can be a way forward. The hardware AI also needs to be distributed, large area and flexible/conformable.

As for any AI applications, ensuring transparency and explainability of robot's actions are essential to guarantee accountability of robots' actions. Preventing biases, misuse and ensuring privacy in robotic data handling is critical for social acceptance. These concerns may conflict with usage of control systems that depend on closed-source technologies.

Lastly, if robotics may speed up decarbonization by automatizing battery recycling, solar panel deployment and housing renovation, it is equally important that robot controllers and hardware are sustainable by design. Promoting energy-efficient compute, reusability of data and algorithms, and biodegradable hardware are promising steps in this regard.

**Author Contributions**

A.B. led the writing and editing of the manuscript, and creation of the images. A.A, M.B, W.B, P.C, M.C., R.D., D.K, K.G, Y.N, D.S. contributed to the manuscript's writing. All authors gave final approval for submission.

**Competing Interests**

The authors declare no competing interests